\documentclass[10pt,twocolumn]{article}

\usepackage[margin=0.75in]{geometry}
\usepackage[T1]{fontenc}
\usepackage[utf8]{inputenc}
\usepackage{times}
\usepackage{microtype}
\usepackage{graphicx}
\usepackage{booktabs}
\usepackage{multirow}
\usepackage{amsmath}
\usepackage{amssymb}
\usepackage{enumitem}
\usepackage{hyperref}
\usepackage{xcolor}
\usepackage{subcaption}

\hypersetup{colorlinks=true,linkcolor=blue,citecolor=blue,urlcolor=blue}

\title{Predictive Training with Latent Imagination for Visual Quadruped Navigation}

\author{Yancheng Zhu$^*$, Wanli Ma, Chen Han, Irvin Haozhe Zhan, Bingfeng Qin, Yixin Xu \\
Anker Spatial Perception Lab \\
{\small $^*$Corresponding author}}

\date{}

\begin{document}

\maketitle

\begin{abstract}
Reinforcement-learning navigation policies for legged robots select actions reactively from current observations and short-term memory, with limited capacity to anticipate how moving obstacles will evolve in the near future. In dynamic environments, this reactivity causes the robot to respond too late because collision risk depends on short-horizon scene structure rather than on current obstacle positions alone.
Lightweight predictive supervision applied to the policy's recurrent state during training can encode anticipatory obstacle dynamics without modifying the inference-time controller. We augment a reactive LSTM-SRU navigation backbone with an auxiliary JEPA-style predictor and SIGReg regularization: during training, the predictor supervises the deterministic hidden state to anticipate its own next state; at inference, it is fully discarded, incurring zero additional computational cost.
On simulated and real-world navigation benchmarks with dynamic obstacles, our method substantially improves navigation success while reducing collision rates through the predictive training signal alone, without additional inference-time parameters. Real-robot deployment on a Unitree Go2 demonstrates zero-shot sim-to-real transfer: the controller navigates cluttered indoor and dynamic outdoor environments without fine-tuning, with evasive behavior consistent with the collision reduction observed in simulation.
\end{abstract}

\section{Introduction}

Autonomous local navigation---deciding where to move next from egocentric observations, without a pre-built global map---is a core capability for mobile robots operating in unstructured environments. For legged robots in particular, this capability is central to applications such as search-and-rescue, industrial inspection, and logistics in complex terrain. As deployment environments become more dynamic, however, a fundamental limitation of learned navigation policies becomes apparent: most are designed to react to the current state of the world, with limited capacity to anticipate how that state will change.

The difficulty is not obstacle detection but anticipation. Classical planners such as the Dynamic Window Approach~\cite{fox1997dwa} and Vector Field Histogram~\cite{borenstein1991vfh} select actions from instantaneous geometry. Learned policies---including reinforcement-learning navigation~\cite{xie2023drlvo}, diffusion-based controllers~\cite{cai2025navdp}, and large-scale goal-conditioned models~\cite{shah2023vint,sridhar2023nomad}---improve generalization but remain fundamentally reactive: they respond to obstacles that are already in a dangerous configuration rather than to where those obstacles are heading. In dynamic environments, this is a structural limitation, because collision risk depends on short-horizon obstacle motion rather than on current positions alone.

World-model methods~\cite{shanks2025dreamernav,bar2025nwm} address this gap by learning latent dynamics that support prediction or planning during inference. These approaches demonstrate that predictive training substantially improves navigation under uncertainty. However, they couple prediction to generative objectives---reconstruction, rollout, or imagination---that must remain active at deployment, imposing computational overhead that constrains real-world applicability on platforms with limited onboard resources.

This work investigates a more constrained setting: whether predictive supervision applied exclusively during training, with the predictive component fully discarded at inference, can measurably improve a reactive policy's performance in dynamic environments. We augment a recurrent navigation backbone with a lightweight auxiliary predictive branch that supervises the policy's hidden state to anticipate its own future, drawing on joint-embedding predictive architectures~\cite{assran2023jepa,lecun2022path} and a regularization term that prevents representational collapse~\cite{leworldmodel_archive}. The deployed controller is structurally identical to a purely reactive policy, incurring no additional inference cost. Experiments on simulated static and dynamic obstacle navigation benchmarks demonstrate that this training-time predictive signal substantially improves navigation success and reduces collisions, with gains attributable specifically to the predictive supervision rather than to additional model capacity or training budget. The trained policy transfers zero-shot to a physical Unitree Go2 quadruped without fine-tuning or domain adaptation.

Our contributions are: (1) a training-time predictive supervision framework for reactive recurrent navigation policies, which improves anticipatory behavior under dynamic obstacles without modifying the inference-time controller; (2) a principled combination of latent prediction and variance-decorrelation regularization that stabilizes hidden-state prediction on deterministic recurrent states; and (3) empirical evidence that predictive training captures local scene dynamics that transfer across task variants and to real hardware, demonstrated through zero-shot sim-to-real deployment on both static and dynamic obstacle scenarios.

\section{Related Work}

\subsection{Learned local navigation}

Classical local planners such as the vector field histogram \cite{borenstein1991vfh} and the dynamic window approach \cite{fox1997dwa} select actions from instantaneous obstacle geometry. Social-force models \cite{helbing1995socialforce} and crowd-aware extensions~\cite{alahi2016social,gupta2018social} introduced interaction-aware motion prediction, but rely on hand-designed interaction rules rather than learned dynamics.

Reinforcement learning recast local navigation as end-to-end policy learning from sensory input. Mapless navigation~\cite{tai2017virtual} demonstrated that learned policies can surpass hand-engineered planners. Recurrent architectures---LSTM~\cite{hochreiter1997lstm}, GRU~\cite{cho2014gru}, and SRU~\cite{yang2025sru}---enabled memory-dependent reasoning under partial observability. The Spatially-Enhanced Recurrent Unit (SRU) specifically addresses spatial registration in egocentric navigation: it aligns hidden memory with the robot's changing viewpoint rather than treating memory as a generic temporal buffer. Large-scale learned policies such as ViNT~\cite{shah2023vint}, NoMaD~\cite{sridhar2023nomad}, DRL-VO~\cite{xie2023drlvo}, and NavDP~\cite{cai2025navdp} further demonstrate the potential of data-driven local controllers. For quadruped robots, SEA-Nav~\cite{chen2026seanav} achieved efficient training for safe navigation in dense clutter through safety-constrained RL.

Despite this progress, these methods share a common limitation: they encode obstacles as spatial features of the current frame, without an explicit mechanism to model how obstacle trajectories evolve over time. Dynamic collision risk is handled through reactive generalization rather than learned anticipation. Our work targets precisely this gap, by adding a lightweight predictive training signal that shapes the recurrent state to encode future-relevant dynamics.

\subsection{World Models and Predictive Learning}

World-model methods learn latent dynamics that support prediction or imagination during training. PlaNet~\cite{hafner2019planet}, DreamerV2~\cite{hafner2020dreamerv2}, DreamerV3~\cite{hafner2023dreamerv3}, and MuZero~\cite{schrittwieser2020muzero} demonstrated that latent predictive training can substantially improve sample efficiency and final performance. DreamerNav~\cite{shanks2025dreamernav} applied this idea to navigation, combining an RSSM world model with multimodal perception. Navigation World Models~\cite{bar2025nwm} push this further, training a billion-parameter diffusion transformer on diverse egocentric video. Structured state-space models such as S4WM~\cite{deng2023s4wm} show that the choice of temporal backbone significantly affects long-range memory quality in model-based RL.

Our work draws on this line but targets a fundamentally different regime. Dreamer-family and NWM-scale methods learn broad generative world models with reconstruction, generation, or planning objectives---designed to support planning at inference. We instead ask a narrower question: whether lightweight predictive supervision alone, without reconstruction or planning, can improve a reactive policy's recurrent representation. This leads us to adopt JEPA-style latent prediction~\cite{assran2023jepa,lecun2022path,assran2025vjepa2}, which targets representational quality through latent agreement rather than observation reconstruction. TD-MPC2~\cite{hansen2024tdmpc2} established that a deterministic latent world model paired with regularization-based collapse prevention can outperform DreamerV3; we extend this principle to navigation, using SIGReg~\cite{leworldmodel_archive} as the anti-collapse mechanism for our deterministic recurrent state. Unlike contrastive predictive coding~\cite{oord2018cpc} or non-contrastive self-supervised objectives~\cite{grill2020byol,chen2021simsiam}, our predictive target is the policy's own recurrent state rather than a projected embedding, making the learned dynamics directly useful for the downstream control task.

\section{Method}

\subsection{Problem Setting}
\label{sec:obs}

We consider visual quadruped navigation in obstacle-rich environments under partial observability. The robot must reach a goal position while avoiding both static and dynamic obstacles, using only onboard depth sensing, proprioception, and a local obstacle representation. This is fundamentally a local navigation problem: no global map is assumed, and the policy must make decisions from egocentric observations that change as the robot moves.

The navigation task is formulated as goal-conditioned reinforcement learning. At each timestep $t$, the agent receives an observation
\begin{equation}
o_t = \{p_t, v_t, d_t\},
\end{equation}
where $p_t \in \mathbb{R}^{16}$ denotes proprioceptive and goal-related features, $v_t$ denotes depth-derived visual features, and $d_t \in \mathbb{R}^{132}$ denotes an explicit obstacle representation encoding up to 12 nearby obstacles with 11 features each (relative position, relative velocity, size, orientation, distance, and collision-threat descriptor). In simulation, $d_t$ is extracted from ground-truth physics state with noise injection for sim-to-real robustness; on the real robot, approximate obstacle features are obtained via depth-based YOLO detection and tracking. The policy outputs a high-level velocity command
\begin{equation}
a_t \in \mathbb{R}^{3} = (v_x, v_y, \omega),
\end{equation}
which is filtered and executed by a frozen low-level locomotion controller on the quadruped. This hierarchical separation is deliberate: the high-level policy focuses on SE(2) navigation decisions, while legged locomotion is handled by a pretrained controller that does not participate in navigation learning.

Three challenges motivate the method design. First, the robot observes the world from a changing egocentric viewpoint; past observations must be implicitly re-aligned to the current body frame to remain useful, a spatial registration problem that standard recurrent memory does not automatically solve \cite{yang2025sru}. Second, obstacle relevance is state-dependent: an obstacle on a collision course deserves more representational weight than one far away and moving away. Third, in dynamic environments, safe behavior depends on short-horizon scene evolution, not only the current scene---a purely reactive policy has no mechanism to represent how the local scene is about to change.

\begin{figure*}[t]
    \centering
    \includegraphics[width=\linewidth]{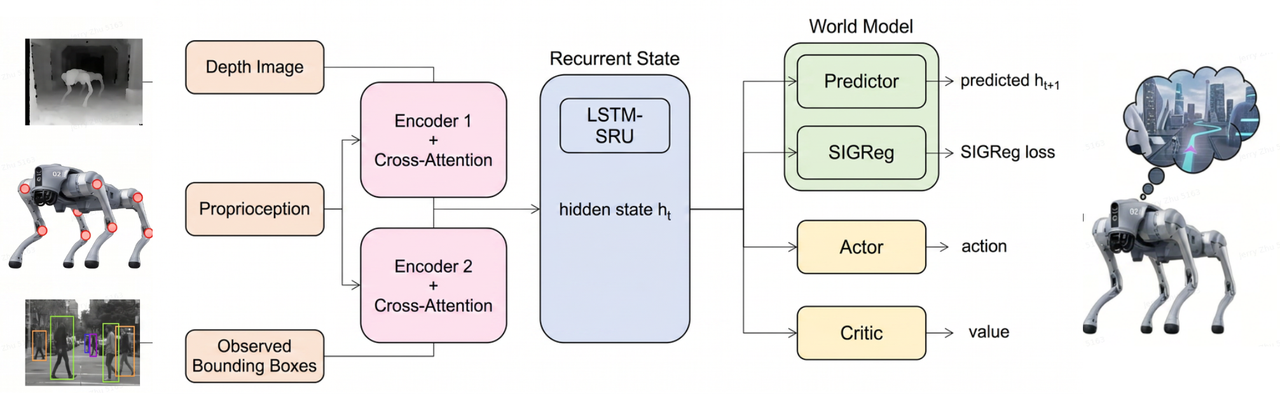}
    \caption{Model architecture of the SRU-WM framework. The reactive backbone (black) fuses depth, proprioception, and obstacle observations through robot-conditioned attention into an LSTM-SRU navigation memory, from which actor and critic heads produce actions and value estimates. The predictive branch (blue) adds training-time latent prediction $\hat{h}_{t+1} = g_\theta(h_t, a_t)$ with SIGReg regularization on $h_t$; the branch is discarded at inference.}
    \label{fig:architecture_overview}
\end{figure*}

Our framework addresses these challenges through a two-part design. The reactive navigation backbone handles the spatial memory and selective-attention challenges: it fuses depth, proprioception, and obstacle observations through robot-conditioned attention into a compact latent embedding, processed by an LSTM-SRU recurrent memory to form the navigation state from which actor and critic heads produce actions and value estimates. The predictive branch targets the anticipation challenge by adding training-time latent supervision that pressures the recurrent state to encode future-relevant scene dynamics; the branch is discarded at inference (Figure~\ref{fig:architecture_overview}).

\subsection{Obstacle Attention and Recurrent Memory}

The obstacle branch encodes local scene structure for the navigation policy. Each of the up to 12 nearest obstacles (represented by $d_{t,i} \in \mathbb{R}^{11}$, as defined in Section~\ref{sec:obs}) is encoded independently by a shared MLP $\phi_{obs}$ into a fixed-dimensional embedding $e^{obs}_i = \phi_{obs}(d_{t,i})$.

A naive approach would concatenate these embeddings into a flat feature vector. However, obstacle relevance is not uniform: an obstacle on a collision course ahead deserves far more representational weight than one far to the side and moving away. To capture this state-dependence, we use a robot-conditioned cross-attention module where the robot's proprioceptive state $p_t$ serves as the query and the obstacle embeddings $\{e^{obs}_i\}$ form the keys and values, producing a threat-aware scene summary $c_t$.
This contrasts with self-attention among obstacles, which models obstacle-obstacle interactions but not robot-conditioned threat.

The fused observation $z_t$ concatenates the visual features, proprioceptive state, and obstacle summary, then feeds into the recurrent backbone:
\begin{equation}
z_t = [\phi_{img}(v_t) ; p_t ; c_t], \qquad h_t = f_{\text{SRU}}(z_t, h_{t-1}),
\label{eq:fused_obs}
\end{equation}
where $\phi_{img}$ is a convolutional depth encoder mapping the $64{\times}40$ depth image to a compact visual embedding.

We adopt the LSTM-SRU variant \cite{yang2025sru} as our recurrent backbone. SRU stands for Spatially-Enhanced Recurrent Unit: it augments the standard LSTM or GRU with a spatial transformation operation that enables the hidden state to align observations across changing egocentric perspectives --- a critical capability for mapless navigation where the robot must register what it sees now against what it saw moments ago from a different viewpoint. The core idea is to modulate the candidate hidden state with a spatial modulation signal $s_t$ computed from odometry:
\begin{equation}
g_t = \tanh\left(s_t \odot \left(W_{xg} x_t + W_{hg}(r_t \odot h_{t-1}) + b_g\right)\right),
\end{equation}
where $\odot$ is element-wise multiplication, $r_t$ is the GRU-style reset gate, and $s_t = f_{\text{odom}}(v_t, \omega_t)$ encodes the robot's linear and angular velocity at step $t$. The spatial signal $s_t$ geometrically transforms the candidate state to account for the viewpoint change between $t-1$ and $t$, so that the hidden memory remains spatially aligned with the environment rather than drifting under perspective shifts. The LSTM variant we employ integrates this spatial modulation into the LSTM cell update:

\begin{align}
r_t &= \sigma(W_r [h_{t-1}, x_t] + b_r), \\
i_t &= \sigma(W_i [h_{t-1}, x_t] + b_i), \\
f_t &= \sigma(W_f [h_{t-1}, x_t] + b_f), \\
\tilde{c}_t &= g_t \quad \text{(spatially-modulated candidate)}, \\
c_t &= f_t \odot c_{t-1} + i_t \odot \tilde{c}_t,\\
h_t &= o_t \odot \tanh(c_t),
\end{align}
where $r_t$ is a reset gate that selectively suppresses prior hidden state before computing the spatial candidate, $i_t$, $f_t$, $o_t$ are the input, forget, and output gates, and $g_t$ replaces the standard LSTM candidate $\tanh(W_c [h_{t-1}, x_t] + b_c)$ with the odometry-modulated variant. This mechanism is particularly effective for navigation: as the robot turns, $s_t$ rotates the candidate hidden state to compensate, preventing the perspective change from washing out previously stored spatial information. Actor and critic heads operate directly on $h_t$:
\begin{equation}
a_t = \pi(h_t), \qquad v_t = V(h_t).
\end{equation}

\begin{figure}[t]
    \centering
    \includegraphics[width=0.8\linewidth]{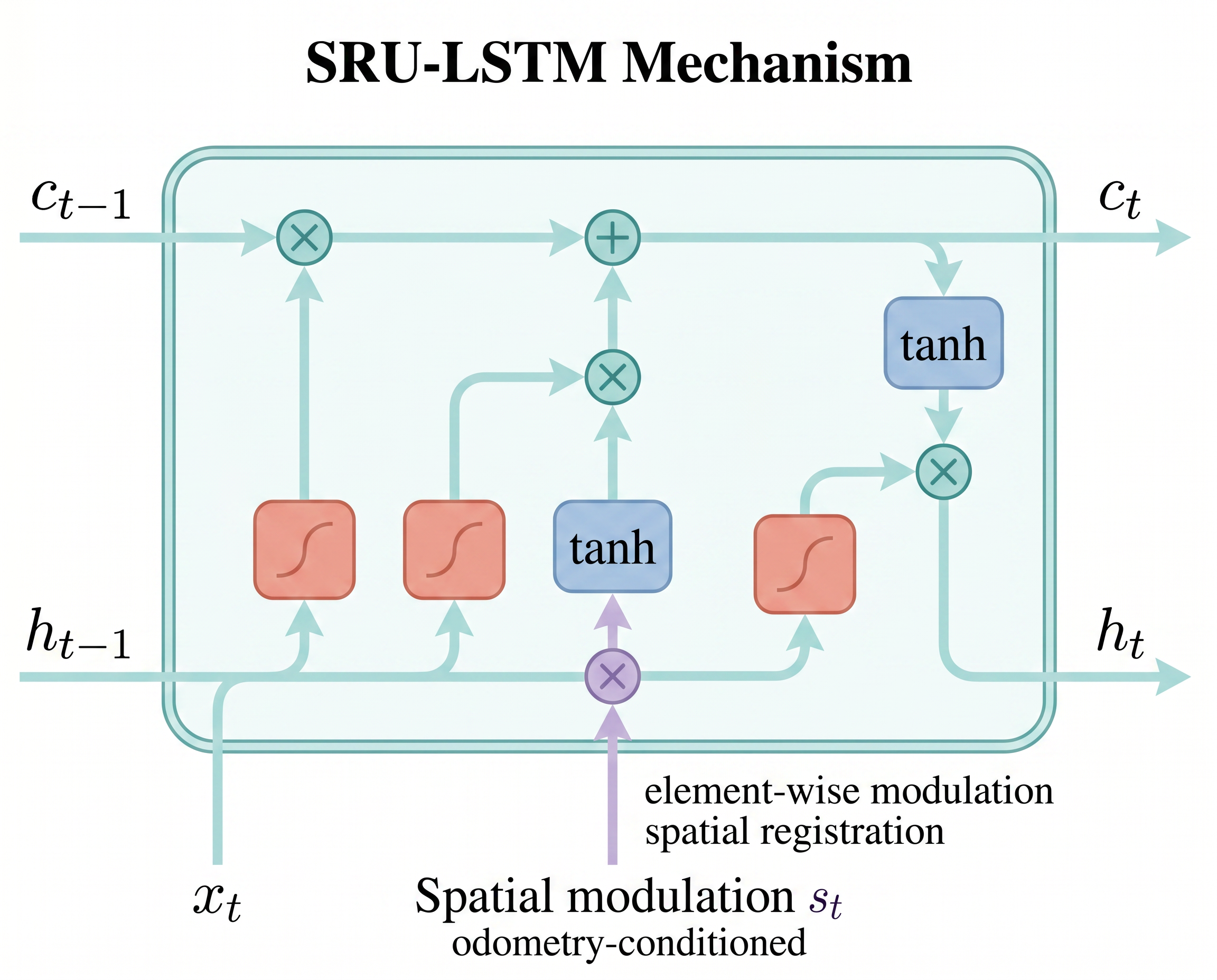}
    \caption{LSTM-SRU recurrent cell with spatial modulation. The odometry-derived spatial signal $s_t$ modulates the candidate hidden state $\tilde{c}_t$, enabling the memory to remain spatially aligned with the environment as the robot's viewpoint changes.}
    \label{fig:lstm_sru}
\end{figure}

The LSTM-SRU cell structure is illustrated in Figure~\ref{fig:lstm_sru}.

\subsection{Predictive Branch}
\label{sec:predictive_branch}

Anticipation under dynamics cannot be solved by reactive memory alone. We address it through an auxiliary predictive branch, active only during training, that pressures the recurrent state to encode future-relevant scene dynamics. Unlike full world models (Dreamer) or online planning methods (TD-MPC2), which learn generative or planning-capable dynamics for inference-time use, our predictive branch serves only as a training signal and is discarded at deployment.

The 512-dimensional $h_t$ is the sole latent representation. A lightweight MLP predictor $g_{\theta}$ learns the transition $h_t \mapsto h_{t+1}$:

\begin{equation}
\hat{h}_{t+1} = g_{\theta}(h_t, a_t).
\end{equation}

Three complementary gradient signals converge on the same LSTM-SRU backbone:

\begin{itemize}[leftmargin=*]
    \item \textbf{RL gradient:} $L_{\text{RL}}$ flows through the actor head into $h_t$, then back to the SRU and encoder. This drives task-level performance---reaching goals and avoiding collisions.
    \item \textbf{Prediction gradient:} $L_{\text{pred}}$ flows from the predictor through $h_t$ into the SRU, pressuring the hidden state to encode future-relevant dynamics. The actor reads the same $h_t$ that is being optimized for prediction, so predictive gradients directly shape the representation used for control.
    \item \textbf{SIGReg gradient:} $L_{\text{sigreg}}$ flows into both the SRU and the encoder, enforcing variance and decorrelation on $h_t$ to prevent representational collapse.
\end{itemize}

Two gradient stop points are critical for stable training. First, the observation embedding $z_t$ is detached from the predictor: this prevents $L_{\text{pred}}$ from corrupting the encoder's visual features, which already receive sufficient training signals from $L_{\text{RL}}$ and $L_{\text{sigreg}}$. Second, the prediction target $h_{t+1}$ is stop-gradiented ($\text{sg}(h_{t+1})$): this prevents the trivial solution where the predictor collapses all targets toward a constant, which would simultaneously minimize $L_{\text{pred}}$ and destroy the temporal structure the actor requires.

The one-step training objective combines all three losses:

\begin{align}
L_{\text{pred}} &= \frac{1}{d}\| g_{\theta}(h_t, a_t) - \text{sg}(h_{t+1}) \|_2^2, \\
L_{\text{total}} &= L_{\text{RL}} + \alpha_{\text{pred}}L_{\text{pred}} + \alpha_{\text{sigreg}}L_{\text{sigreg}},
\end{align}
where $d$ is the hidden state dimension and the $\frac{1}{d}$ normalization keeps $L_{\text{pred}}$ scale-invariant to the choice of hidden state size.

The three gradients are complementary rather than conflicting. $L_{\text{RL}}$ tells $h_t$ what action leads to high reward; $L_{\text{pred}}$ tells $h_t$ what information predicts the next state. These are naturally aligned: a representation that encodes how the local scene will evolve is also more informative for selecting safe actions, since collision risk depends on short-horizon scene dynamics rather than instantaneous obstacle positions. The SIGReg term ensures this alignment does not degenerate into collapse.

\paragraph{Risk of representational collapse and the role of SIGReg.}
\label{sec:sigreg}
The key risk of predicting in $h$-space without a stochastic bottleneck is representational collapse: if all $h_t$ converge to a constant, the predictor trivially achieves zero loss. SIGReg~\cite{leworldmodel_archive} addresses this by enforcing a structured empirical distribution on $h_t$:

\begin{equation}
L_{\text{sigreg}} = \frac{1}{d}\sum_i (1 - \sigma_i)^2 + \frac{1}{d}\sum_{i \neq j} C_{ij}^2,
\end{equation}

where $\sigma_i$ is the standard deviation of dimension $i$ across the batch, and $C_{ij}$ is the off-diagonal entry of the batch correlation matrix (covariance normalized by the product of standard deviations), so that $C_{ij} \in [-1, 1]$ and the penalty is dimensionless. The variance term encourages unit variance per dimension; the covariance term penalizes correlation across dimensions. This serves the same functional role as the KL divergence in DreamerV3~\cite{hafner2023dreamerv3}---preventing collapse---but through a simpler, non-generative mechanism that does not require learning a posterior network or sampling. The design is related to VICReg~\cite{bardes2022vicreg} and Barlow Twins~\cite{zbontar2021barlow} but applied to deterministic recurrent states rather than projected embeddings. In practice, the SIGReg loss stabilizes at $\approx$2.06 throughout training (Section~\ref{sec:ablation}), confirming that the representation maintains healthy variance.

\paragraph{Remark on multi-step rollout.}
The one-step predictor $g_\theta$ can in principle be applied recursively to produce short-horizon latent trajectories: $\hat{h}_{t+k} = g_{\theta}(\hat{h}_{t+k-1}, a_{t+k-1})$ for $k = 1, \ldots, H$. This extension is architecturally compatible with the current framework but is not included in the reported experiments; all results in Section~5 use one-step prediction only. Multi-step rollout is left for future investigation.

One practical implementation detail is that predicting directly in $h$-space requires normalizing the hidden-state scale, which can be unstable during early training. We apply LayerNorm to $h_t$ before it enters the predictor, normalizing the variance to approximately 1.0 and decoupling the predictor's learning dynamics from the evolving scale of the recurrent state.

\section{Training}

\subsection{Training and Inference}

\begin{figure}[t]
    \centering
    \includegraphics[width=\linewidth]{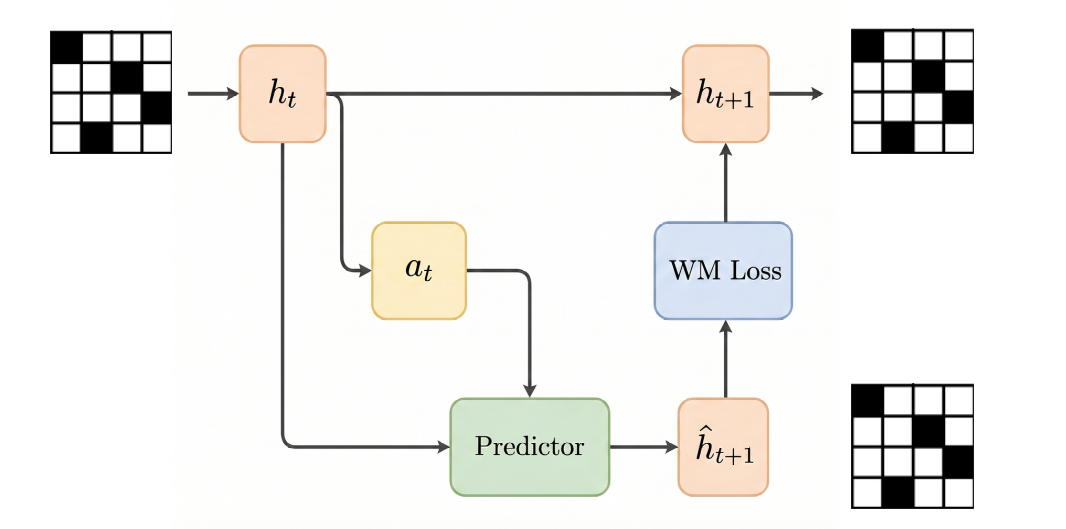}
    \caption{World-model training process. The LSTM-SRU backbone produces latent states; the predictive branch performs one-step prediction and optional rollout; auxiliary losses supervise hidden dynamics. The lower path is training-only.}
    \label{fig:wm_training_process}
\end{figure}

The high-level navigation policy is optimized with Mirror Descent Policy Optimization (MDPO)~\cite{tomar2022mdpo}, a trust-region RL algorithm that subsumes PPO~\cite{schulman2017ppo} as a special case, while the predictive branch trains through its own auxiliary losses. An important asymmetry separates actor and critic: the actor receives only deployable observations $(p_t, v_t, d_t)$, whereas the critic additionally receives privileged information---true obstacle states and terrain structure---unavailable on the real robot. This asymmetric design allows the value function to learn from richer supervision without contaminating the actor's observation space, following the teacher--student pattern common in sim-to-real RL \cite{miki2022wild}. The policy is trained on navigation episodes. At deployment, the predictive branch is discarded entirely. The deployed controller consists only of the reactive backbone: observation encoding, obstacle attention, LSTM-SRU memory, and the actor head. No imagination rollout, planning loop, or world-model query occurs at runtime. Predictive training shapes what the controller has learned, not what it does at each step (Figure~\ref{fig:wm_training_process}).

\subsection{Reward Design}

The reward function is a multi-term shaped objective designed to encourage goal-directed, smooth, and collision-averse behavior without collapsing into trivial low-speed policies. Table~\ref{tab:reward} lists all components.

\begin{table}[t]
\centering
\caption{Reward function components.}
\label{tab:reward}
\small
\resizebox{\columnwidth}{!}{\begin{tabular}{lrl}
\toprule
Term & Weight & Form \\
\midrule
reach\_goal\_soft & $+0.5$ & $\sigma^+(d_{xy}, k{=}2.5)$ \\
reach\_goal\_tight & $+2.5$ & $\sigma^+(d_{xy}, k{=}0.25) \cdot \mathbf{1}_{d<1\text{m}}$ \\
motion\_preference & $+0.01$ & toward-goal velocity component \\
lateral\_movement & $-0.1$ & $|v_y|$ \\
action\_rate\_l1 & $-0.1$ & $\|a_t - a_{t-1}\|_1$ \\
joint\_acc\_l2 & $-10^{-7}$ & $\|\ddot{q}\|_2^2$ \\
episode\_termination & $-50.0$ & $\mathbf{1}_{\text{failed}}$ \\
\bottomrule
\end{tabular}}
\end{table}

Goal reaching uses a two-stage soft positive kernel $\sigma^+(d, k)$ with sharpness $k$: the soft term ($k{=}2.5$) provides a broad gradient basin from any distance, while the tight term ($k{=}0.25$) activates only within 1\,m to sharpen final approach. Motion preference gives a small positive signal for velocity aligned with the goal direction, preventing the policy from learning to stand still as a collision-avoidance strategy. Smoothness penalties discourage lateral drift, jerky actions, and excessive joint acceleration---all critical for sim-to-real transfer. The failure penalty ($-50$) is large enough to dominate short episodes, making collision avoidance a strong implicit objective. We distinguish training termination from evaluation success: episodes terminate when the robot reaches within 0.5\,m of the goal (sticky counter with 3\,s grace) to prevent reward farming, while evaluation success uses a 2\,m radius consistent with the NavRL and NavDP evaluation protocol \cite{xie2023drlvo,cai2025navdp}.

Episode termination distinguishes truncations from true failures. Timeout (60\,s), goal reached, and terrain fall are bootstrapped truncations; base contact (rollover) and large pitch angle ($>$40$^\circ$) are true failures with $V(\text{terminal})=0$. The sticky goal counter increments only forward and never resets, preventing reward farming by repeatedly entering and leaving the goal zone.

The policy is trained with MDPO, which in our setting maintains two actor--critic networks regularized toward each other through symmetric KL coupling. Matrix-valued parameters use the Muon optimizer~\cite{jordan2024muon}; all remaining parameters use AdamW~\cite{loshchilov2019adamw}. The predictive loss and SIGReg regularization are weighted at $\alpha_{\text{pred}} = 0.03$ and $\alpha_{\text{sigreg}} = 0.003$, calibrated to shape the representation without disrupting policy optimization.

\subsection{Implementation}

\begin{figure}[t]
    \centering
    \begin{subfigure}[b]{0.48\linewidth}
        \includegraphics[width=\linewidth,height=4cm]{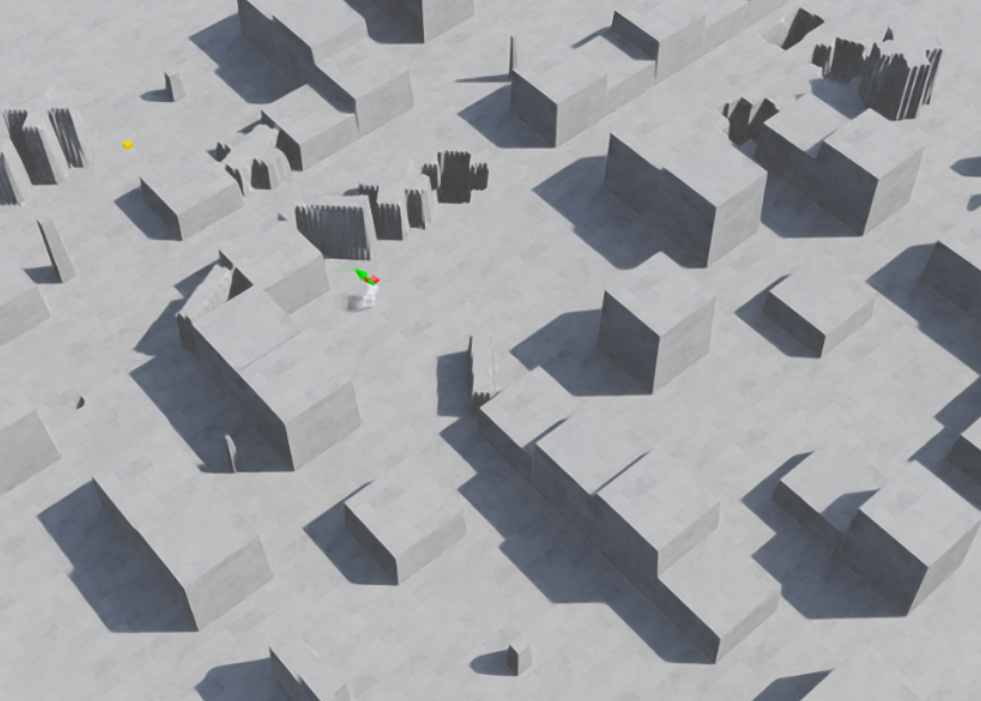}
        \subcaption{Maze terrain.}
        \label{fig:env_maze}
    \end{subfigure}
    \begin{subfigure}[b]{0.48\linewidth}
        \includegraphics[width=\linewidth,height=4cm]{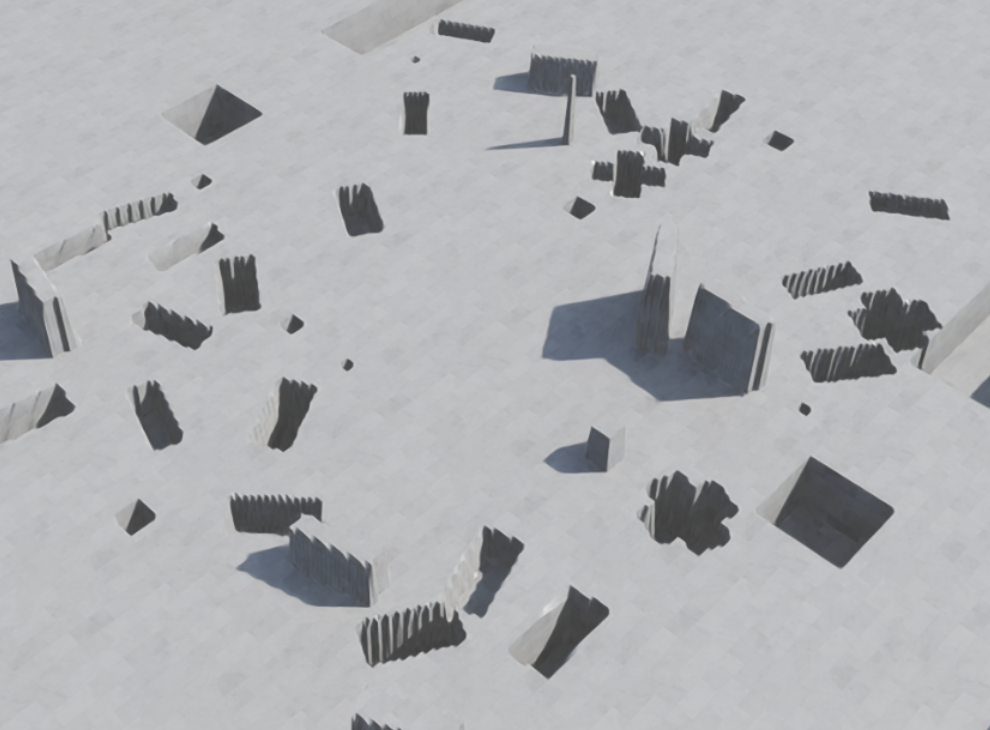}
        \subcaption{Pillar terrain.}
        \label{fig:env_pillar}
    \end{subfigure}

    \vspace{4pt}
    
    \begin{subfigure}[b]{0.48\linewidth}
        \includegraphics[width=\linewidth,height=4cm]{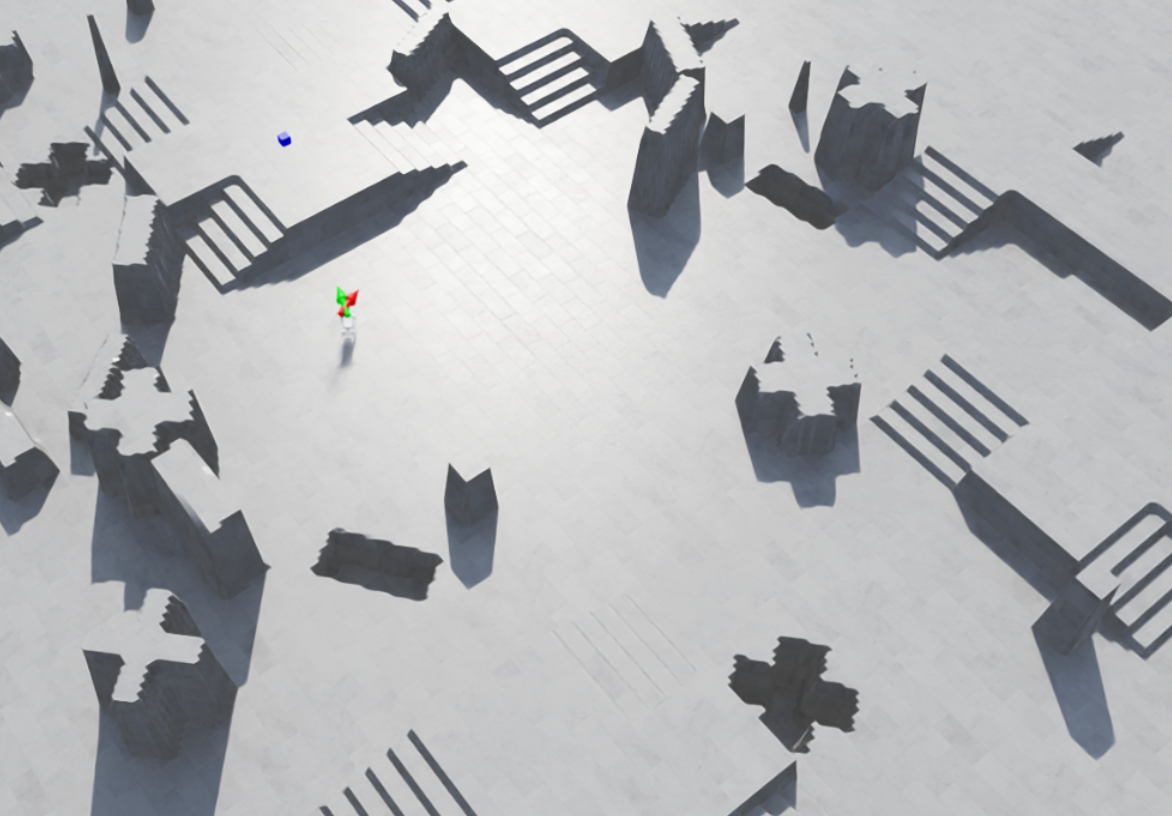}
        \subcaption{Stairs terrain.}
        \label{fig:env_stairs}
    \end{subfigure}
    \begin{subfigure}[b]{0.48\linewidth}
        \includegraphics[width=\linewidth,height=4cm]{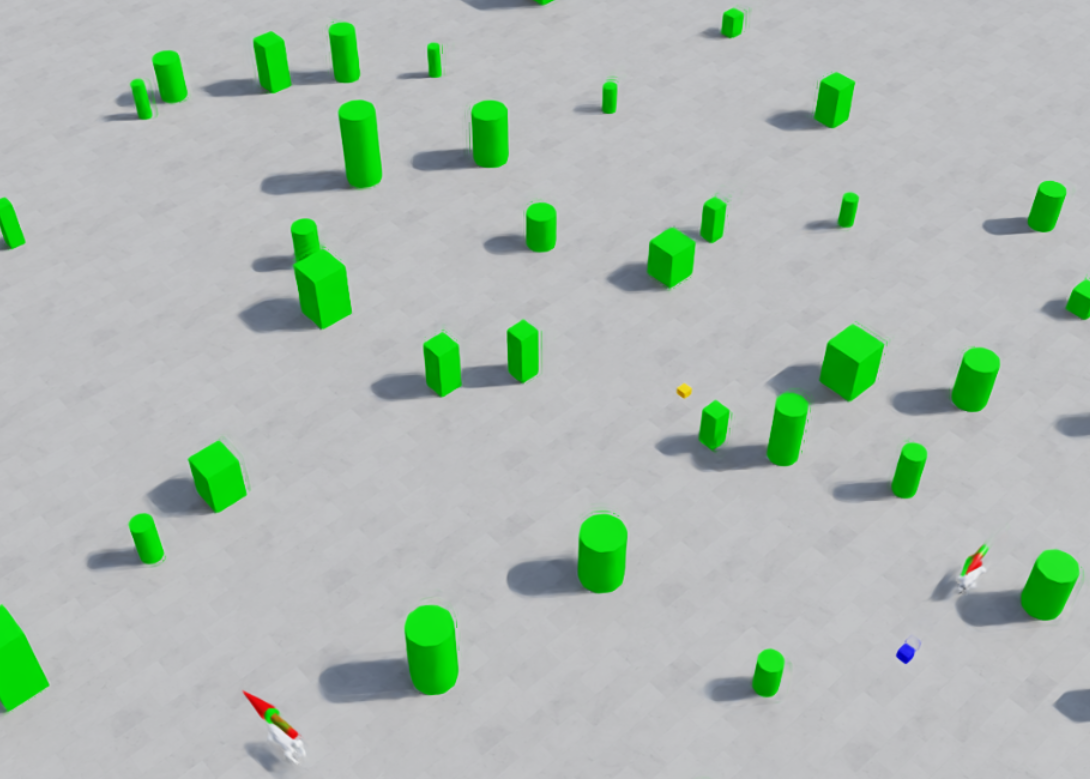}
        \subcaption{Dynamic obstacle arena.}
        \label{fig:env_dynamic}
    \end{subfigure}
    \caption{Simulated training environments spanning diverse terrain types: maze, pillar field, stairs, and dynamic obstacle arena.}
    \label{fig:env_overview}
\end{figure}

The environment spans diverse terrain types including mazes, pillar fields, stairs, and dynamic-obstacle arenas (Figure~\ref{fig:env_overview}). Depth images ($64{\times}40$) are encoded via pretrained VAE features and fused with obstacle states through cross-attention. Dynamic obstacle speed follows a linear curriculum, scaling from half to full speed over the initial training phase to prevent the policy from facing the hardest dynamic scenarios prematurely. Domain randomization is applied to friction coefficients, action scaling, low-pass filter parameters, and external forces for sim-to-real transfer. Depth input receives disparity-based stereo noise to prevent reliance on idealized sensor readings.

Training uses GPU-accelerated Isaac Sim~4.5 / Isaac Lab~2.1.1 with PhysX simulation. The system operates at three frequencies: physics at 200\,Hz, the low-level locomotion controller at 50\,Hz, and the high-level navigation policy at 5\,Hz. At each navigation step, the policy outputs $(v_x, v_y, \omega)$. This command passes through tanh compression, per-axis policy scaling (with randomized gain during training), and a low-pass filter (exponential smoothing, $\alpha$ randomized per axis). The filtered command is executed by the pretrained Go2 locomotion policy at 50\,Hz. This post-processing pipeline smooths policy output, absorbs sim-to-real mismatch, and prevents the locomotion controller from receiving discontinuous commands. For real-robot deployment, depth input is provided by an Intel RealSense camera processed to match the $64{\times}40$ simulation resolution; no additional planning, mapping, or post-processing is introduced beyond training. All models are trained for 20,000 iterations across 4,096 parallel environments on 4 NVIDIA RTX 4090D GPUs, with a total wall-clock time of approximately 3 days.

\section{Experiments}

Our experiments address two questions: (1) does the predictive-training framework improve navigation, and which components drive the gain (internal ablation); and (2) how does the resulting policy compare to published navigation methods (external benchmark)? We evaluate on two task variants---single-goal navigation (Nav) and navigation with dynamic obstacles (DynObs)---each under matched environment settings.

\subsection{Setup}

We compare against two published methods: NavRL~\cite{xie2023drlvo}, a LiDAR-based RL navigation policy, and NavDP~\cite{cai2025navdp}, an RGB-D diffusion-policy navigation method. Both are evaluated using their official pre-trained checkpoints, adapted to run in our Isaac Lab simulation environment with matched terrain, obstacle density, and episode length. Because NavRL and NavDP were designed for different sensor configurations and action spaces, we treat this as a system-level comparison rather than a controlled ablation: the goal is to situate our method relative to published state-of-the-art performance, not to isolate the contribution of individual components. NavDP's high timeout rate on the Nav task reflects its diffusion-policy inference latency under our real-time control budget, not a deficiency of the method in its native setting. We also include SRU (our backbone without the predictive branch) as a direct controlled baseline that shares our observation space, reward, and training protocol. For all tasks, we report success rate (SR), collision rate (CR), and timeout rate (TO), where success is defined as reaching within 2\,m of the goal.

For internal ablation, we evaluate four configurations on the same SRU backbone with depth observations and obstacle attention: (1) \textit{SRU (no WM)}---the reactive baseline without predictive training; (2) \textit{SRU-WM (no SIGReg)}---predictive branch without the regularizer; (3) \textit{SRU-WM (Transformer)}---predictive branch with a Transformer predictor; and (4) \textit{SRU-WM (MLP)}---our full configuration with an MLP predictor and SIGReg. All variants share identical training budgets on Isaac Lab 2.1.1 with Unitree Go2. Metrics are averaged over the final 10\% of training steps.

\subsection{Ablation Study}
\label{sec:ablation}

Table~\ref{tab:ablation} reports benchmark evaluation metrics for the architecture variants on the Nav task. Episodes are classified by final robot-goal distance: success (within 2\,m), collision (early termination outside 2\,m), and timeout (episode length $\ge$ 85\% of max).

\begin{table}[t]
\centering
\caption{Ablation study on the Nav task (SR / CR / TO: success, collision, timeout rate; mutually exclusive, sum to 100\%).}
\label{tab:ablation}
\footnotesize
\begin{tabular}{lrrr}
\toprule
Variant & SR (\%) & CR (\%) & TO (\%) \\
\midrule
SRU (no WM)            & 90.7 &  5.6 & 3.7 \\
SRU-WM (no SIGReg)     & 86.0 & 12.1 & 1.9 \\
SRU-WM (Transformer)   & 95.3 &  4.7 & 0.0 \\
SRU-WM (MLP)          & 94.4 &  5.6 & 0.0 \\
\bottomrule
\end{tabular}
\end{table}

The predictive branch improves SRU~(no~WM) from 90.7\% to 94.4\% success rate while eliminating timeouts, confirming that latent prediction during training shapes a more robust recurrent state. The SIGReg ablation is decisive: removing it drops success rate to 86.0\% and raises collision rate to 12.1\%, both worse than SRU~(no~WM), demonstrating that the regularization term is necessary to prevent representational collapse. Comparing predictor architectures, the Transformer variant achieves 95.3\% SR --- marginally higher but with 2.1M parameters versus 0.4M for the MLP predictor, making the MLP the more practical choice.

\begin{figure}[t]
    \centering
    \begin{subfigure}[b]{\linewidth}
        \includegraphics[width=\linewidth]{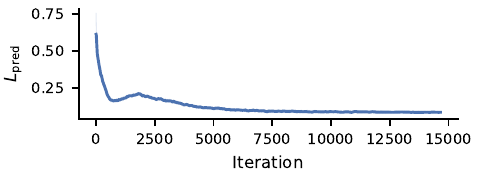}
        \subcaption{Prediction loss $L_{\text{pred}}$.}
        \label{fig:pred_loss}
    \end{subfigure}
    \vspace{2pt}
    \begin{subfigure}[b]{\linewidth}
        \includegraphics[width=\linewidth]{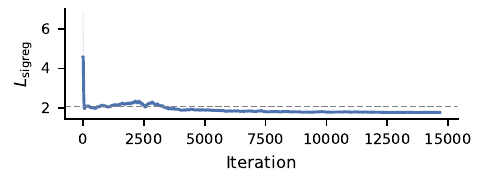}
        \subcaption{SIGReg regularization loss $L_{\text{sigreg}}$.}
        \label{fig:sigreg_loss}
    \end{subfigure}
    \caption{Training loss curves for the predictive branch. Shaded traces are raw values; solid lines are smoothed. Dashed line marks $L_{\text{sigreg}} \approx 2.06$.}
    \label{fig:training_losses}
\end{figure}

Figure~\ref{fig:training_losses} traces the evolution of the two auxiliary losses across training. The prediction loss $L_{\text{pred}}$ decreases steadily from 0.754 at initialization to 0.085 at convergence, indicating that the predictor progressively learns to anticipate the recurrent state's next configuration. The SIGReg loss $L_{\text{sigreg}}$ drops rapidly in the first 20 iterations as the representation escapes initialization, then stabilizes near 2.06 for the remainder of training---confirming that the hidden state maintains healthy variance and does not collapse toward a constant, which would trivially minimize $L_{\text{pred}}$ while destroying the temporal structure the actor requires.

\subsection{Comparison with Published Methods}

The ablation results confirm that the predictive branch and SIGReg regularization together are responsible for the performance gain over the reactive backbone. We now ask whether this gain holds against published navigation baselines operating under the same evaluation protocol, including methods with access to richer sensor modalities.

\begin{figure}[t]
    \centering
    \begin{subfigure}[b]{\linewidth}
        \includegraphics[width=0.95\linewidth]{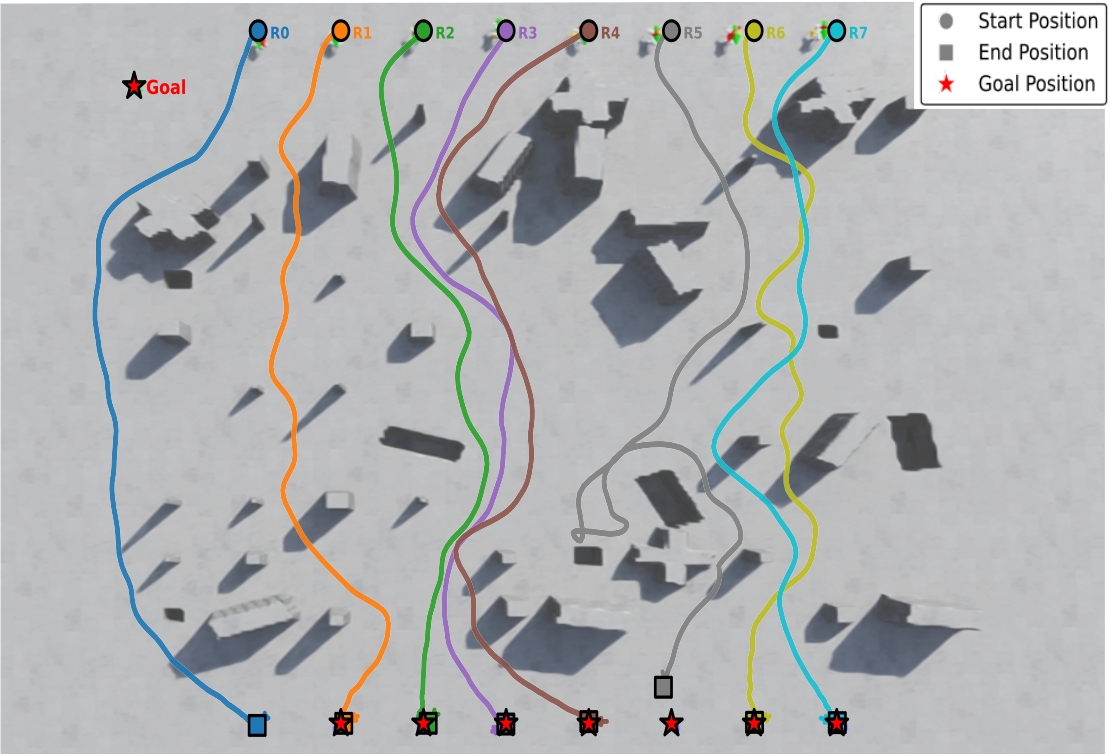}
        \subcaption{SRU (no WM)}
        \label{fig:benchmark_static_sru}
    \end{subfigure}
    \vspace{4pt}
    \begin{subfigure}[b]{\linewidth}
        \includegraphics[width=0.95\linewidth]{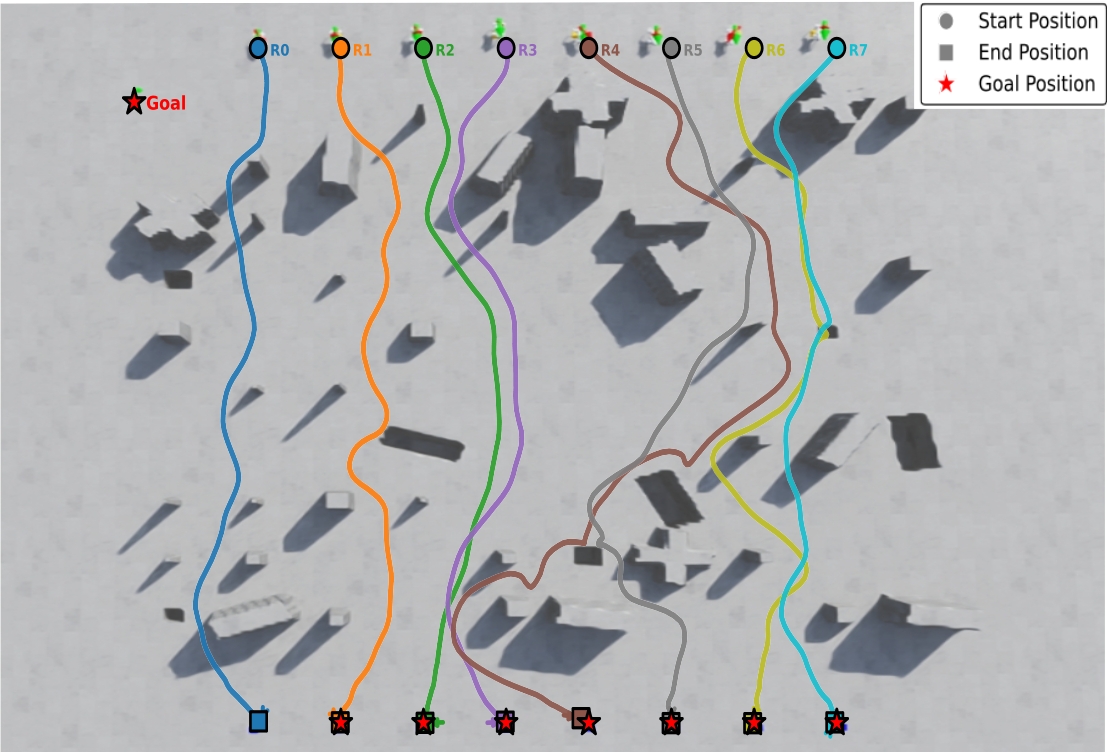}
        \subcaption{SRU-WM (ours)}
        \label{fig:benchmark_static_sruwm}
    \end{subfigure}
    \caption{Qualitative trajectory comparison on static obstacle navigation. SRU-WM maintains straighter paths with fewer hesitation turns compared to the reactive backbone (SRU), reflecting the predictive training improves spatial reasoning through mazes.}
    \label{fig:benchmark_static}
\end{figure}

\begin{figure}[t]
    \centering
    \includegraphics[width=\linewidth]{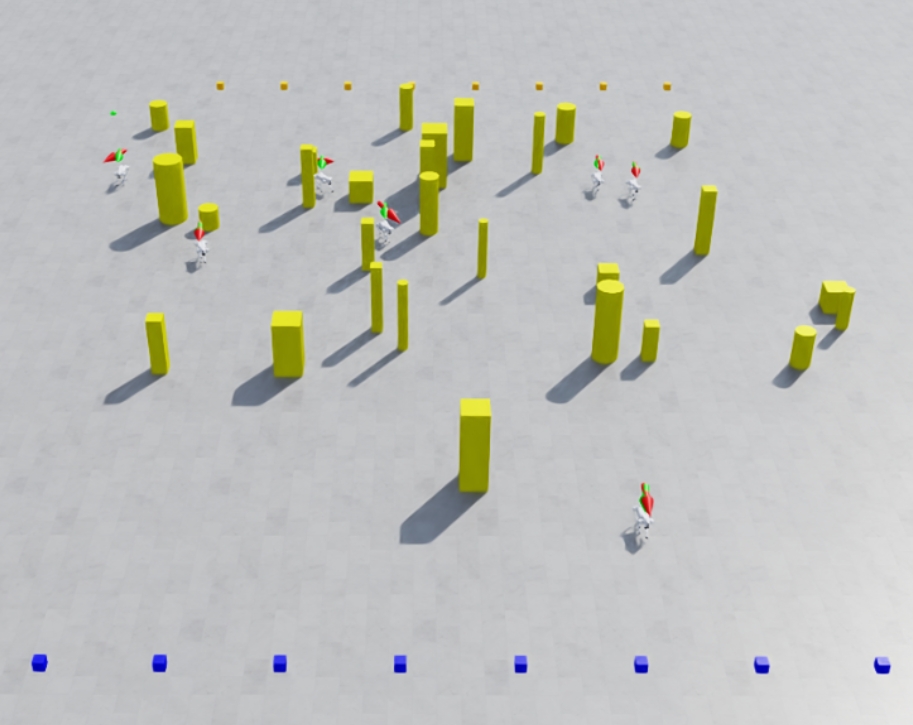}
    \caption{Dynamic obstacle navigation environment. Pedestrians with WANDER, CROSS, and ARC motion patterns traverse the scene at 0.3--0.8\,m/s, creating time-varying collision risk.}
    \label{fig:benchmark_dynobs}
\end{figure}

Tables~\ref{tab:nav}--\ref{tab:dynobs} report comparisons against published baselines on the Nav and DynObs tasks. Each method is evaluated for 107 episodes across 8 parallel environments on identical terrain configurations, with both static obstacles and moving obstacles (WANDER, CROSS, and ARC motion patterns at 0.3--0.8\,m/s for DynObs; Figure~\ref{fig:benchmark_dynobs}).

\begin{table}[t]
\centering
\caption{Navigation goal-reaching (Nav task).}
\label{tab:nav}
\footnotesize
\begin{tabular}{lrrr}
\toprule
Method & SR (\%) & CR (\%) & TO (\%) \\
\midrule
NavDP~\cite{cai2025navdp}   & 11.2 & 14.0 & 74.8 \\
NavRL~\cite{xie2023drlvo}   & 41.5 & 55.7 &  2.8 \\
SRU (no WM)                 & 90.7 &  5.6 &  3.7 \\
SRU-WM (ours)               & 94.4 &  5.6 &  0.0 \\
\bottomrule
\end{tabular}
\end{table}

\begin{table}[t]
\centering
\caption{Dynamic obstacle avoidance (DynObs task).}
\label{tab:dynobs}
\footnotesize
\begin{tabular}{lrrr}
\toprule
Method & SR (\%) & CR (\%) & TO (\%) \\
\midrule
NavDP~\cite{cai2025navdp}   & 19.2 & 59.6 & 21.2 \\
NavRL~\cite{xie2023drlvo}   & 52.3 & 44.9 &  2.8 \\
SRU (no WM)                 & 86.0 & 14.0 &  0.0 \\
SRU-WM (ours)               & 96.2 &  3.8 &  0.0 \\
\bottomrule
\end{tabular}
\end{table}

On the Nav task (Table~\ref{tab:nav}), SRU-WM achieves 94.4\% SR --- $2.3\times$ that of NavRL and $8.4\times$ that of NavDP. The reactive baseline SRU (no WM) already reaches 90.7\%, indicating that the obstacle attention and LSTM-SRU backbone provide a strong foundation; the predictive branch contributes an additional 3.7 percentage points, pushing the policy to near-ceiling performance while eliminating all timeouts.

On the DynObs task (Table~\ref{tab:dynobs}), the margin increases substantially. SRU-WM achieves 96.2\% SR with only 3.8\% collisions; the collision rate is $12\times$ lower than NavRL (44.9\%) and $16\times$ lower than NavDP (59.6\%). NavDP's 59.6\% collision rate confirms that diffusion-policy navigation, trained primarily on static scenes, does not transfer to dynamic environments. NavRL's 44.9\% collision rate, despite using LiDAR, reveals a structural limitation: reactive LiDAR-based policies detect obstacles at the current timestep but cannot anticipate their motion. SRU (no WM) already reduces collisions to 14.0\% through attention-based obstacle conditioning; the predictive branch further reduces this to 3.8\%. The same predictor trained on navigation episodes achieves this reduction on the DynObs task, indicating that the learned recurrent representation encodes short-horizon obstacle motion rather than task-specific heuristics.

\subsection{Real-Robot Transfer}

\begin{figure}[t]
    \centering
    \begin{subfigure}[b]{0.48\linewidth}
        \includegraphics[width=\linewidth]{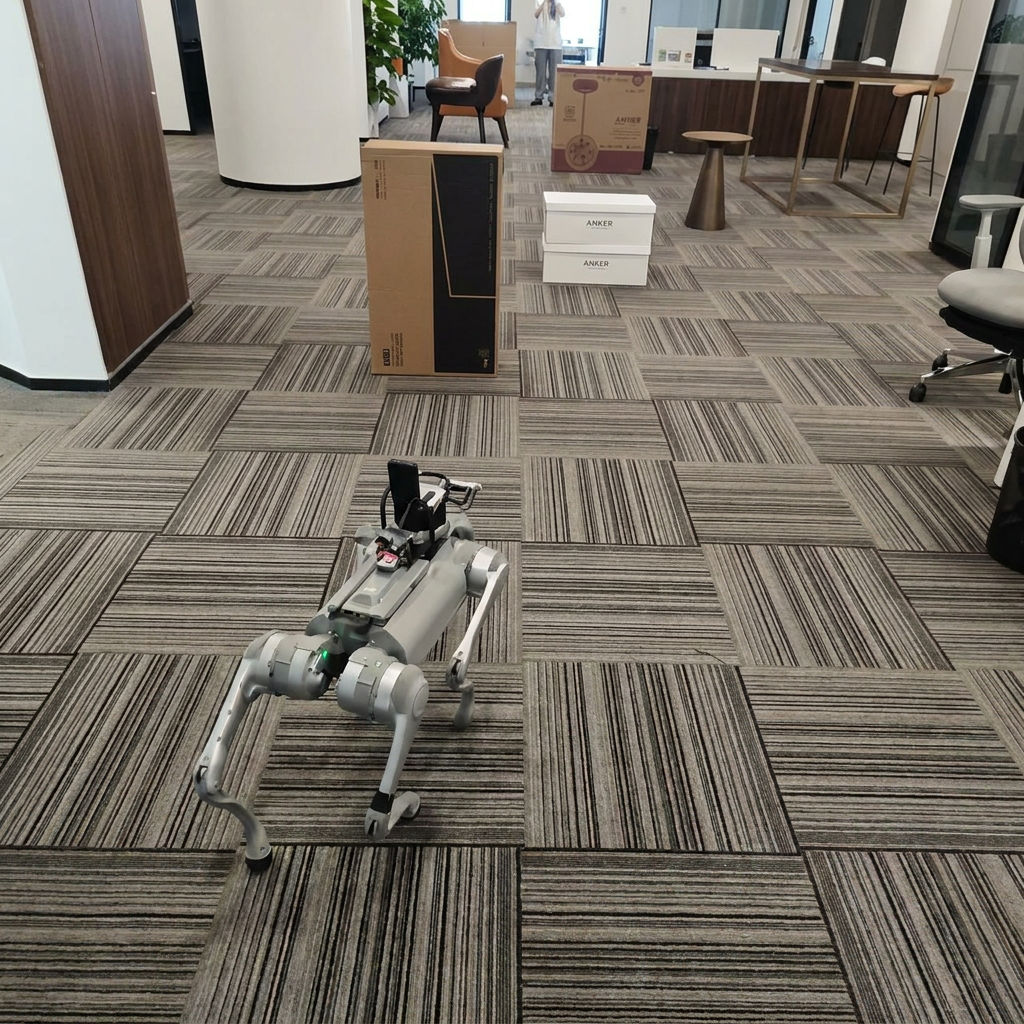}
        \subcaption{$t = 00{:}00$\,s}
    \end{subfigure}
    \hfill
    \begin{subfigure}[b]{0.48\linewidth}
        \includegraphics[width=\linewidth]{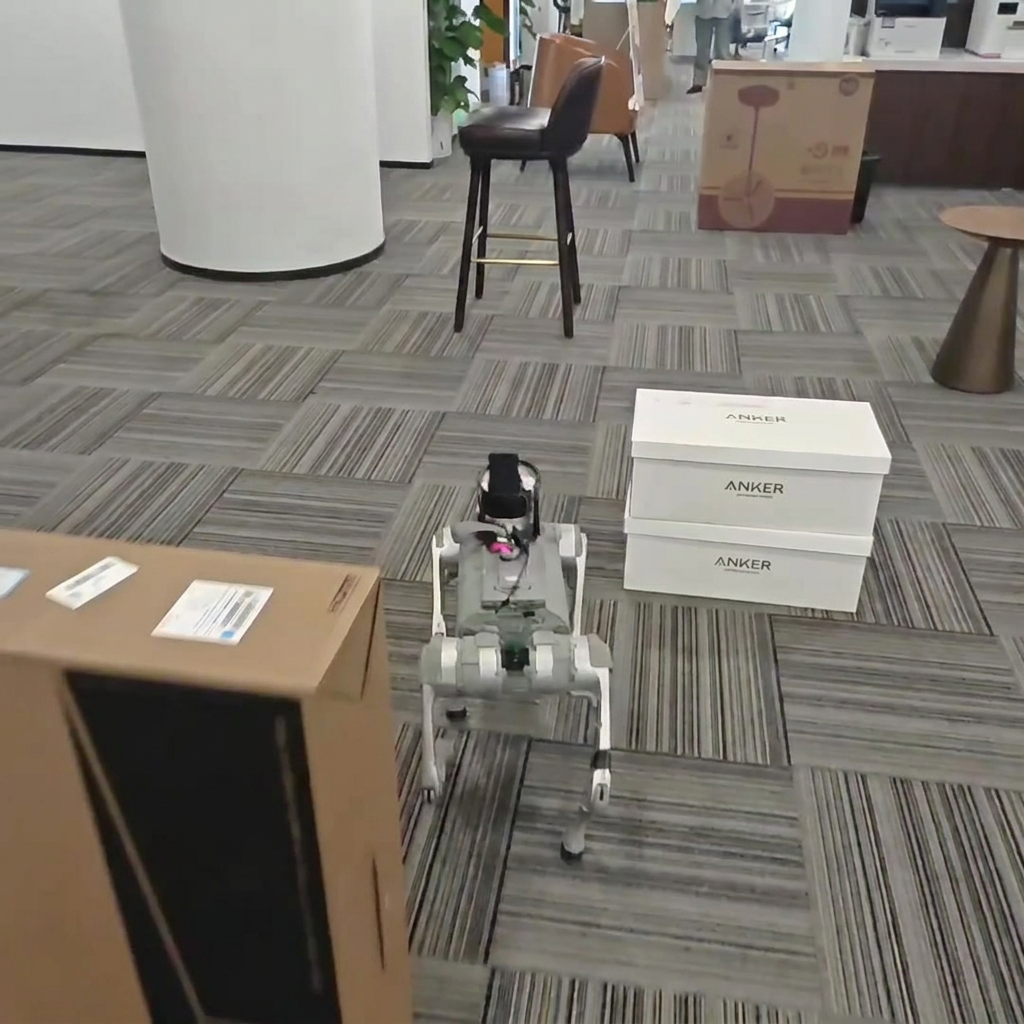}
        \subcaption{$t = 00{:}04$\,s}
    \end{subfigure}
    \vspace{4pt}
    \begin{subfigure}[b]{0.48\linewidth}
        \includegraphics[width=\linewidth]{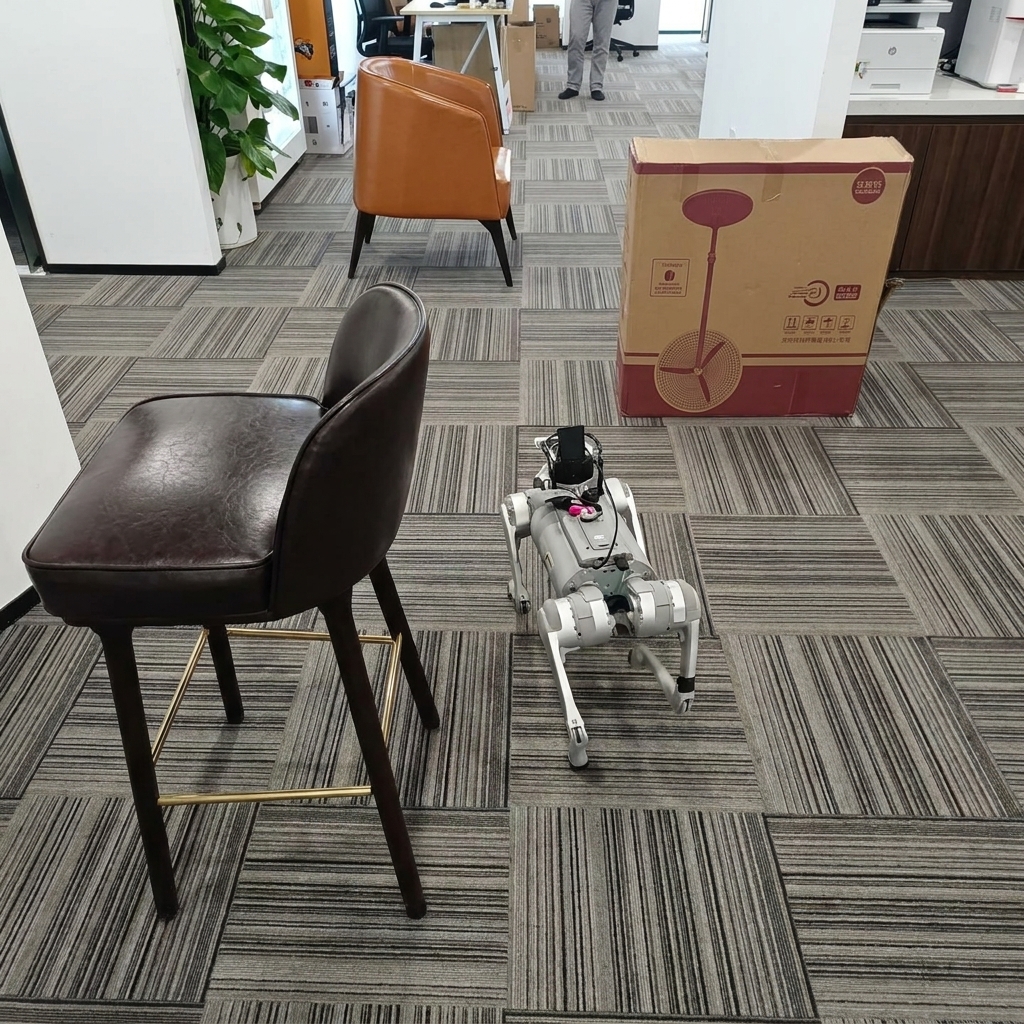}
        \subcaption{$t = 00{:}08$\,s}
    \end{subfigure}
    \hfill
    \begin{subfigure}[b]{0.48\linewidth}
        \includegraphics[width=\linewidth]{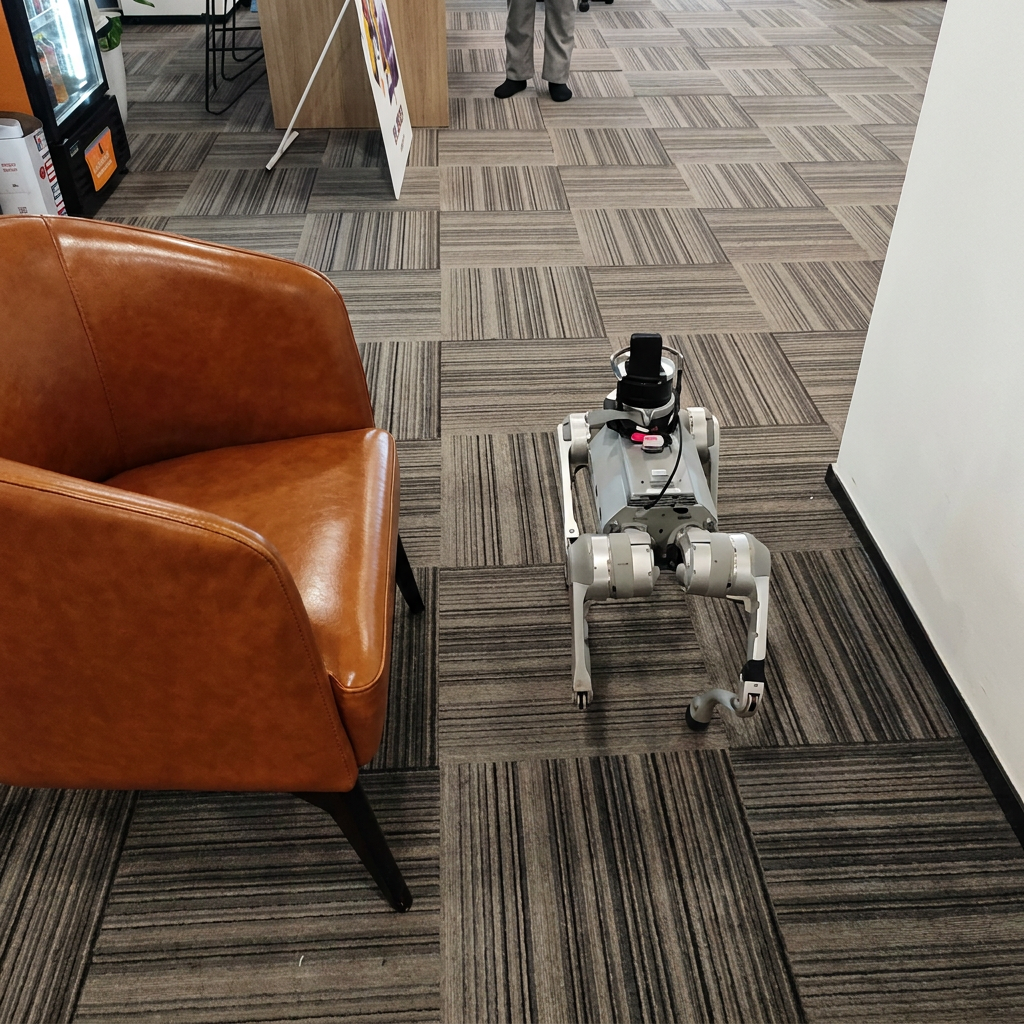}
        \subcaption{$t = 00{:}12$\,s}
    \end{subfigure}
    \caption{Indoor static obstacle avoidance. Four sequential frames show the robot navigating a cluttered environment with fixed obstacles, maintaining consistent heading toward the goal across 12 seconds.}
    \label{fig:real_static}
\end{figure}

\begin{figure}[t]
    \centering
    \begin{subfigure}[b]{\linewidth}
        \includegraphics[width=\linewidth]{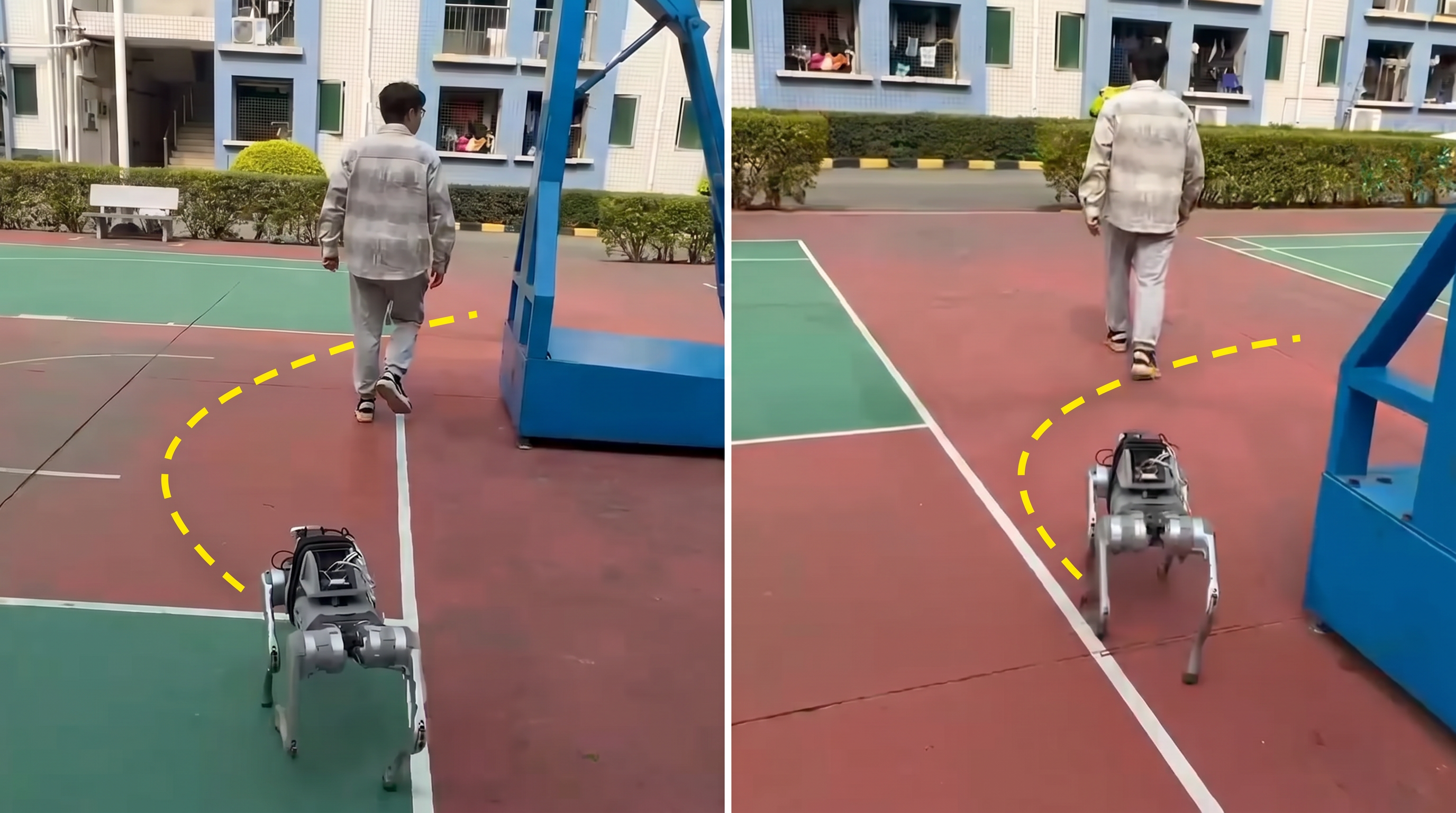}
        \subcaption{Static obstacle avoidance}
    \end{subfigure}
    \vspace{4pt}
    \begin{subfigure}[b]{\linewidth}
        \includegraphics[width=\linewidth]{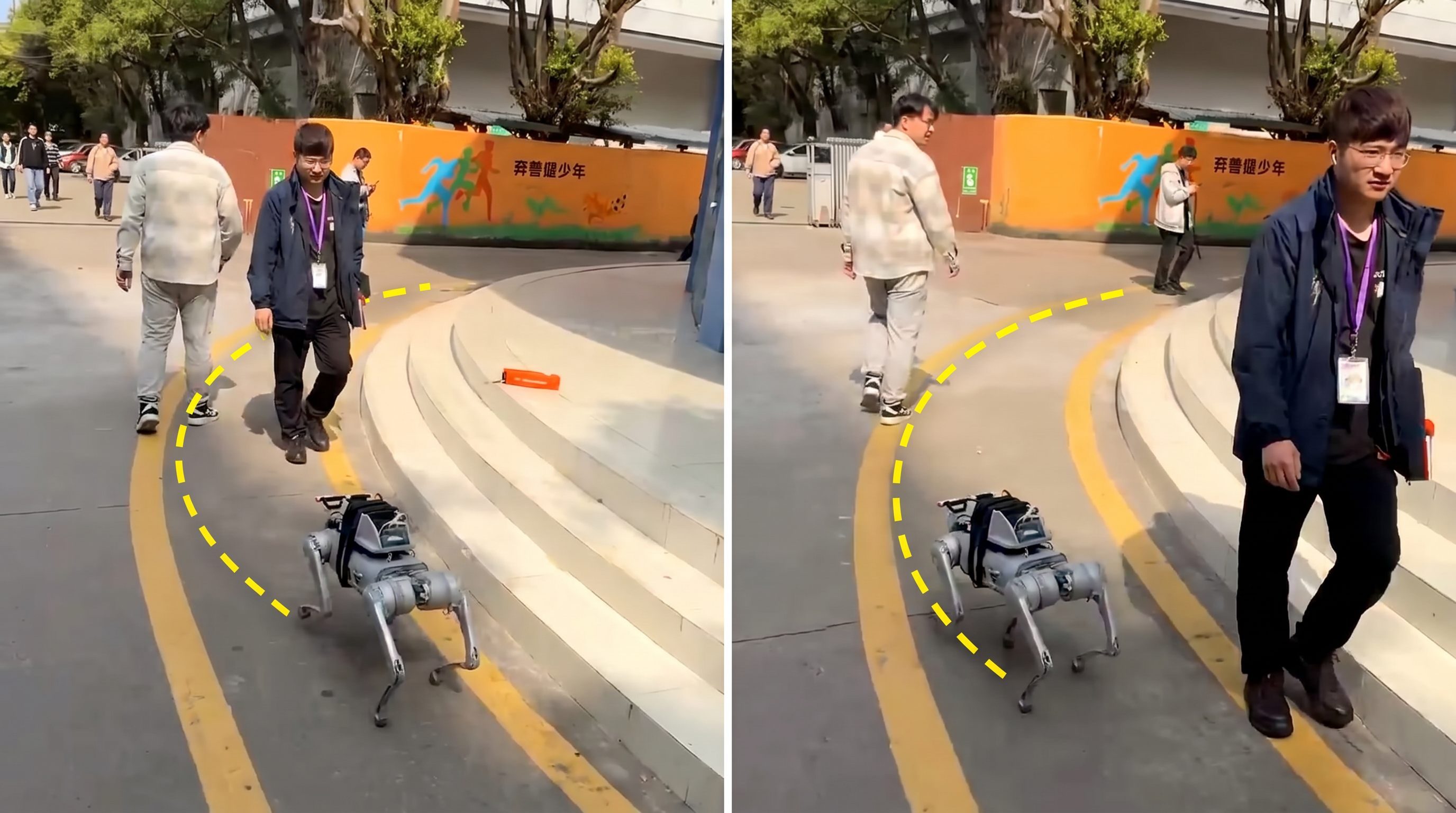}
        \subcaption{Dynamic obstacle avoidance}
    \end{subfigure}
    \caption{Outdoor deployment. Top: static obstacle avoidance in an open outdoor environment. Bottom: dynamic obstacle avoidance against an approaching human; the robot initiates evasive steering while the human is still outside immediate collision range. This behavior is consistent with the simulation-based collision reduction reported in Table~\ref{tab:dynobs}.}
    \label{fig:real_outdoor}
\end{figure}

Our full model was deployed on a Unitree Go2 quadruped equipped with an Intel RealSense D435i depth camera, with all inference running on an onboard NVIDIA Jetson Orin. The navigation policy runs at 5\,Hz; each policy step issues a high-level SE(2) velocity command that the frozen locomotion controller tracks at 50\,Hz. The policy transfers zero-shot: no fine-tuning or domain adaptation is applied, and depth frames from the D435i are resized to match the simulated resolution and passed directly to the trained policy. Figure~\ref{fig:real_static} shows four sequential frames of indoor static obstacle avoidance, where the robot maintains consistent heading through a cluttered environment without collision. Figure~\ref{fig:real_outdoor} shows outdoor deployment: static obstacle navigation (top) and dynamic obstacle avoidance against an approaching human (bottom), where the robot initiates evasive steering while the human is still outside immediate collision range. This behavior is consistent with the simulation-based collision reduction reported in Table~\ref{tab:dynobs}.

Together, these deployments confirm zero-shot sim-to-real transfer across indoor and outdoor settings.

\subsection{Discussion}

The same predictor trained on navigation episodes reduces DynObs collisions from 14.0\% to 3.8\%, indicating that the learned recurrent representation encodes short-horizon obstacle motion rather than task-specific navigation heuristics---the predictive gradients shape what the memory stores, and that structure transfers across task variants.

The source of the gain differs by task. On Nav, the 3.7 percentage-point SR improvement and the elimination of timeouts suggest that predictive supervision produces more stable navigation rather than simply reducing collisions. On DynObs, the larger margin points to a more direct mechanism: pure reactive policies must infer obstacle motion implicitly by integrating velocity across frames in the recurrent state, whereas predictive supervision makes this integration explicit---$L_{\text{pred}}$ directly penalizes hidden states that fail to anticipate the next state, creating a gradient that shapes the recurrent representation toward encoding motion information.

An open question is whether this predictive-training recipe generalizes beyond the quadruped setting---to other robot morphologies, sensor modalities, or environments with longer-horizon dynamics where one-step prediction may be insufficient.

\section{Conclusion}

We have presented a training-time predictive supervision framework for reactive visual quadruped navigation. During training, a JEPA-style auxiliary predictor with SIGReg regularization pressures the policy's recurrent state to encode short-horizon obstacle dynamics; at deployment, the predictor is discarded entirely, leaving the inference architecture unchanged. Predictive gradients reshape what the recurrent memory encodes, and this effect persists at deployment even after the predictor is removed. Experiments confirm measurable gains in both static and dynamic navigation scenarios, with particularly strong improvement in dynamic-obstacle avoidance; zero-shot transfer to a physical Unitree Go2 further validates that the learned dynamics encode generalizable local scene structure rather than simulation-specific artifacts.

Future work includes extending predictive supervision to longer rollout horizons for rare collision-critical events, and decomposing the recurrent state into task-relevant components to enable more structured dynamics learning.

\bibliographystyle{plain}
\bibliography{references}

\end{document}